\pdfoutput=1

\documentclass[dvipsnames,11pt]{article}

\usepackage{EMNLP2022}

\usepackage{times}
\usepackage{latexsym}
\usepackage{graphicx}
\usepackage{bm}

\usepackage[T1]{fontenc}

\usepackage[utf8]{inputenc}

\usepackage{microtype}
\usepackage{multirow}
\usepackage{multicol}
\usepackage{booktabs}

\usepackage{inconsolata}
\usepackage{url}
\usepackage{eurosym}
\usepackage{todonotes}
\usepackage{subcaption}
\usepackage{amssymb}
\usepackage{amsmath}
\usepackage{enumitem}
\usepackage{array}
\usepackage{longtable}
\usepackage{makecell}
\usepackage{xspace}
\usepackage{xcolor,colortbl}
\usepackage{tcolorbox}
\usepackage{arydshln}
\usepackage{adjustbox}
\newcommand{\dline}{\hdashline[0.5pt/1pt]}
\usepackage{tikz}
\usepackage[tikz]{bclogo}
\usepackage{pgfplots}
\pgfplotsset{width=1.0\columnwidth}

\usepackage{makecell}
\usepackage{pifont}
\usepackage{bbm}
\usepackage{tablefootnote}
\usepackage{soul}

\newcommand{\figref}[1]{Figure~\ref{fig:#1}}
\newcommand{\secref}[1]{\S\,\ref{sec:#1}}
\newcommand{\tabref}[1]{Table~\ref{tab:#1}}
\usepackage{graphicx}

\newcommand{\tagop}{\textsc{TagOp}\xspace}
\newcommand{\tapas}{\textsc{TaPas}\xspace}

\newcommand{\ours}{\textsc{PoEt}\xspace}
\newcommand{\ourssql}{\textsc{PoEt}-SQL\xspace}

\newcommand{\oursroberta}{\textsc{PoEt}-SQL$_{\rm {\color{JungleGreen} {\tt{RoBERTa}}}}$\xspace}

\newcommand{\oursbart}{\textsc{PoEt}-SQL$_{\rm {\color{Magenta} {\tt{BART}}}}$\xspace}
\newcommand{\oursbartmath}{\textsc{PoEt}-Math\xspace}

\newcommand{\ourstfive}{\textsc{PoEt}-SQL$_{\rm {\color{Orange} {\tt{T5}}}}$\xspace}

\newcommand{\oursmath}{\textsc{PoEt}-Math\xspace}

\newcommand{\PreserveBackslash}[1]{\let\temp=\\#1\let\\=\temp}

\newcommand\fone{F\textsubscript{1}}
\newcommand*{\affaddr}[1]{#1}
\newcommand*{\affmark}[1][*]{\textsuperscript{#1}}
\newcommand*{\email}[1]{\tt{\small #1}}

\newcommand{\reducedstrut}{\vrule width 0pt height .95\ht\strutbox depth .95\dp\strutbox\relax}
\newcommand{\context}[1]{
\begingroup
\setlength{\fboxsep}{0pt}%
\colorbox[HTML]{DDE9D3}{\reducedstrut\textbf{#1}}
\endgroup
}
\newcommand{\sentence}[1]{
\begingroup
\setlength{\fboxsep}{0pt}%
\colorbox[HTML]{FBF2CD}{\reducedstrut\textbf{#1}}
\endgroup
}

\definecolor{black}{RGB}{0,0,0}
\newcommand{\finding}[1]{
	\begin{bclogo}[couleur=black!05, epBord=0, logo=\bclampe, tailleOndu=3, barre=none,sousTitre={\em #1}]{} 
	\end{bclogo}
}

\title{Reasoning Like Program Executors}

\author{Xinyu Pi\affmark[$\lozenge$]{\thanks{~~The first two authors contributed equally. Work done during internship at Microsoft Research Asia.}}~~, Qian Liu\affmark[\S]$^*$, Bei Chen\affmark[\textdagger], \textbf{Morteza Ziyadi}\,\affmark[$\heartsuit$], Zeqi Lin\affmark[\textdagger] \\
\textbf{Qiang Fu}\affmark[\textdagger], \textbf{Yan Gao}\affmark[\textdagger], \textbf{Jian-Guang Lou}\affmark[\textdagger], \textbf{Weizhu Chen}\affmark[$\heartsuit$]\\
\affaddr{\affmark[$\lozenge$]University of Illinois Urbana-Champaign, Urbana, USA}; \affaddr{\affmark[\S]Sea AI Lab, Singapore}\\
\affaddr{\affmark[\textdagger]Microsoft Research Asia, Beijing, China}; \affaddr{\affmark[$\heartsuit$]Microsoft Azure AI, Redmond, WA, USA}\\
\email{xinyupi2@illinois.edu;} \email{liuqian@sea.com;}\\
\email{\{beichen, morteza.ziyadi, zeqi.lin, qifu, yan.gao, jlou, wzchen\}@microsoft.com}}

\begin{document}
\maketitle
\begin{abstract}
Reasoning over natural language is a long-standing goal for the research community.
However, studies have shown that existing language models are inadequate in reasoning.
To address the issue, we present \ours, a novel reasoning pre-training paradigm.
Through pre-training language models with programs and their execution results, \ours empowers language models to harvest the reasoning knowledge possessed by program executors via a data-driven approach.
\ours is conceptually simple and can be instantiated by different kinds of program executors.
In this paper, we showcase two simple instances \ours-Math and \ours-Logic, in addition to a complex instance, \ours-SQL.
Experimental results on six benchmarks demonstrate that \ours can significantly boost model performance in natural language reasoning, such as numerical reasoning, logical reasoning, and multi-hop reasoning.
\ours opens a new gate on reasoning-enhancement pre-training, and we hope our analysis would shed light on the future research of reasoning like program executors.

\end{abstract}

\section{Introduction}

Recent breakthroughs in pre-training illustrate the power of pre-trained \textbf{L}anguage \textbf{M}odels (LM) on a wide range of \textbf{N}atural \textbf{L}anguage (NL) tasks.
Pre-training on self-supervised tasks, such as masked language modeling~\citep{devlin-etal-2019-bert, he2020deberta} using large amounts of NL sentences, boosts the language understanding of models by a large margin~\citep{wang-etal-2018-glue}.
However, existing pre-training paradigms have primarily focused on language modeling and paid little attention to advanced \textit{reasoning} capabilities (\tabref{discrete_reasoning}).
As a result, though reaching near-human performance on several tasks, pre-trained LMs are still far behind expectations in reasoning-required scenarios~\citep{Jack2022scaling}, such as numerical reasoning~\citep{wallace-etal-2019-nlp,ravichander-etal-2019-equate} and logical reasoning~\citep{yu2020reclor,ijcai2020-501}.

\begin{figure}[t]
    \centering
    \includegraphics[width=1.0\columnwidth]{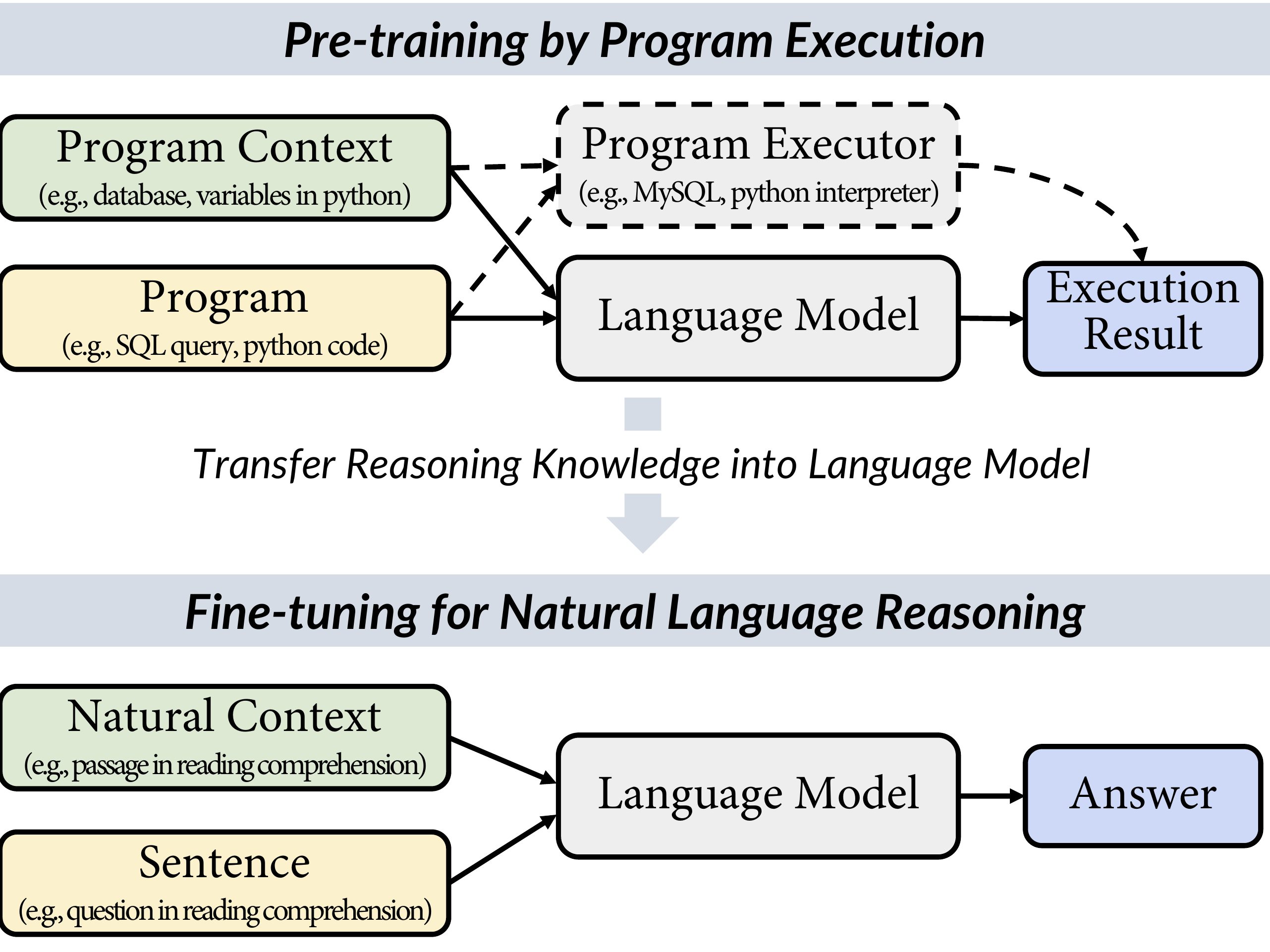}
    \caption{Given a program context and a program as input, \ours pre-trains LMs to output the execution result.
    After fine-tuning on downstream tasks, \ours can boost LMs on reasoning-required scenarios.
    Explanations about program context, program, program executor and execution result can be found in~\secref{reasoning}.
    More examples of natural context and sentence are in~\tabref{discrete_reasoning}.}
    \label{fig:overview}
\end{figure}

\begin{table*}[t]
    \small
    \centering
    \scalebox{0.88}{
    \begin{tabular}{cp{9.8cm}p{1.8cm}p{3.2cm}}
    \toprule
    \textbf{Type} & \textbf{Example} & \textbf{Dataset} & \textbf{Task}\\
    \toprule
    Numerical & \sentence{Question:} What is the difference in casualty numbers between Bavarian and Austrian? \context{Passage:} \texttt{[DOC]} The popular uprising included large areas of $\dots$ & DROP~\citep{dua-etal-2019-drop} &
    Reading Comprehension (RC)
    \\
    \midrule
    Logical & \sentence{Conclusion:} One employee supervises another who gets more salary than himself. \context{Fact:} \texttt{[DOC]} David, Jack and Mark are colleagues in a company. David supervises Jack, and Jack supervises Mark. David gets more $\dots$ & LogiQA \citep{ijcai2020-501} & Reading Comprehension (RC) \\
    \midrule
    \multirow{2}{*}{Multi-hop} &  \sentence{Question:} At which university does the biographer of John Clare teach English Literature? \context{Passage:} \texttt{[DOC]} John Clare : John Clare was an English poet $\dots$ \texttt{[DOC]} CMS College Kottayam : The CMS College is one $\dots$  & HotpotQA \citep{yang-etal-2018-hotpotqa} & Reading Comprehension (RC) \\
    \midrule
    Hybrid & \sentence{Question:} What was the percentage change in gaming between 2018 and 2019? \context{Context:} \texttt{[TAB]} Server products and cloud services | $32,622$ | $26,129$ $\dots$ \texttt{[DOC]} Our commercial cloud revenue, which includes Office $\dots$ & TAT-QA \citep{zhu-etal-2021-tat} & Question\,Answering (QA) \\
    \midrule
    Quantitative & \sentence{Hypothesis:} Teva earns \$7 billion a year. \context{Premise:} After the deal closes, Teva will generate sales of about \$7 billion a year, the company said. & EQUATE \citep{ravichander-etal-2019-equate} & Natural Language Inference (NLI) \\
    \bottomrule
    \end{tabular}
    }
    \caption{The demonstration of five representative reasoning types. Listed are the types, the example questions, the representative dataset, and their corresponding tasks. \texttt{[DOC]} and \texttt{[TAB]} indicates the start of a passage and a semi-structured table respectively.
    Here we regard \sentence{Question}, \sentence{Conclusion} and \sentence{Hypothesis} as \textit{sentence}, and \context{Passage}, \context{Fact}, \context{Context} and \context{Premise} as \textit{natural context} in~\figref{overview}.}
    \label{tab:discrete_reasoning}
\end{table*}

To alleviate the deficiency, reconciling NL understanding in LMs and reasoning in symbolic representations, i.e., neuro-symbolic reasoning, has been a major area of interest~\citep{DBLP:journals/corr/abs-1711-03902,DBLP:journals/aiopen/ZhangCZKD21}.
With a hybrid architecture, i.e., symbolic reasoners attached to LMs, neural-symbolic reasoning shines in a variety of reasoning tasks~\citep{Chen2020Neural,DBLP:conf/aaai/TuHW0HZ20,wolfson-etal-2020-break}.
However, the reasoning mechanism remains in the symbolic reasoner and is not internalized into LMs, making it difficult to reuse the reasoning mechanism on unseen tasks.
Meanwhile, neural models are notorious for their reliance on correlations among concrete tokens of a representation system and are usually assumed to be hard to grasp abstract rules of a symbolic reasoner~\citep{helwe2021reasoning, sinha-etal-2021-unnatural}.
This drives us to explore whether symbolic reasoning can be internalized by language models and, especially,

\noindent\fbox{\begin{minipage}{\dimexpr\columnwidth-2\fboxsep-2\fboxrule\relax}
\centering
Can \textbf{neural} language models advance reasoning abilities by imitating \textbf{symbolic} reasoners?
\end{minipage}}
\hspace{2em}

Motivated by this, we conceive a new pre-training paradigm, \ours (\textbf{P}r\textbf{o}gram \textbf{E}xecu\textbf{t}or), to investigate the learnability of language models from  symbolic reasoning and transferrability across distinct representation systems.
As illustrated in~\figref{overview}, with a \textit{program} (e.g., SQL query) and its \textit{program context} (e.g., database) as input, the model receives automatic supervision from an established \textit{program executor} (e.g., MySQL) and learns to produce correct \textit{execution result}.
By imitating program execution procedures, we believe LMs could potentially learn the reasoning knowledge that humans adopted to create the associated program executor and tackle NL sentences with the learned reasoning capability.
This reveals the key hypothesis of \ours: \textit{program executors are crystallized knowledge of formal reasoning, and such knowledge can be grasped by language models and transferred to NL reasoning via pre-training}.
In other words, pre-training over natural language might be a contingent condition for LMs to have better reasoning capabilities over natural language.

This contingency assumption of NL brings \ours another great merit in data quality: while it is typically difficult to obtain large amounts of clean natural language sentences containing clear evidence of reasoning, synthesized programs can be made arbitrarily complicated but readily available on any scale, thanks to the artificial and compositional nature of programming languages.
These merits greatly facilitate the construction of high-quality corpora, addressing most of the unresolved shortcomings in previous reasoning-enhancement pre-training.
In other words, \ours differs from existing pre-training paradigms relying on noisy NL data.
In summary, our contribution is three-fold:
\begin{itemize}
    \item We propose \ours, a new pre-training paradigm for boosting the reasoning capabilities of language models by imitating program executors. Along with this paradigm, we present three exemplary across-program \ours instantiations for various reasoning capabilities.
    \item We show with quantitative experiments that the reasoning ability our models obtains from \ours pre-training is transferable to broader natural language scenarios. On six reasoning-focused downstream tasks, \ours enables general-purpose language models to achieve competitive performance.
    \item We carry out comprehensive analytical studies, summarize insightful open questions, and provide insights for future work. We hope these insights would shed light on more research on reasoning like program executors\footnote{The code is available at \url{https://github.com/microsoft/ContextualSP}}.
\end{itemize}

\section{Related Work}

\begin{figure*}[t]
    \centering
    \includegraphics[width=0.62\textwidth]{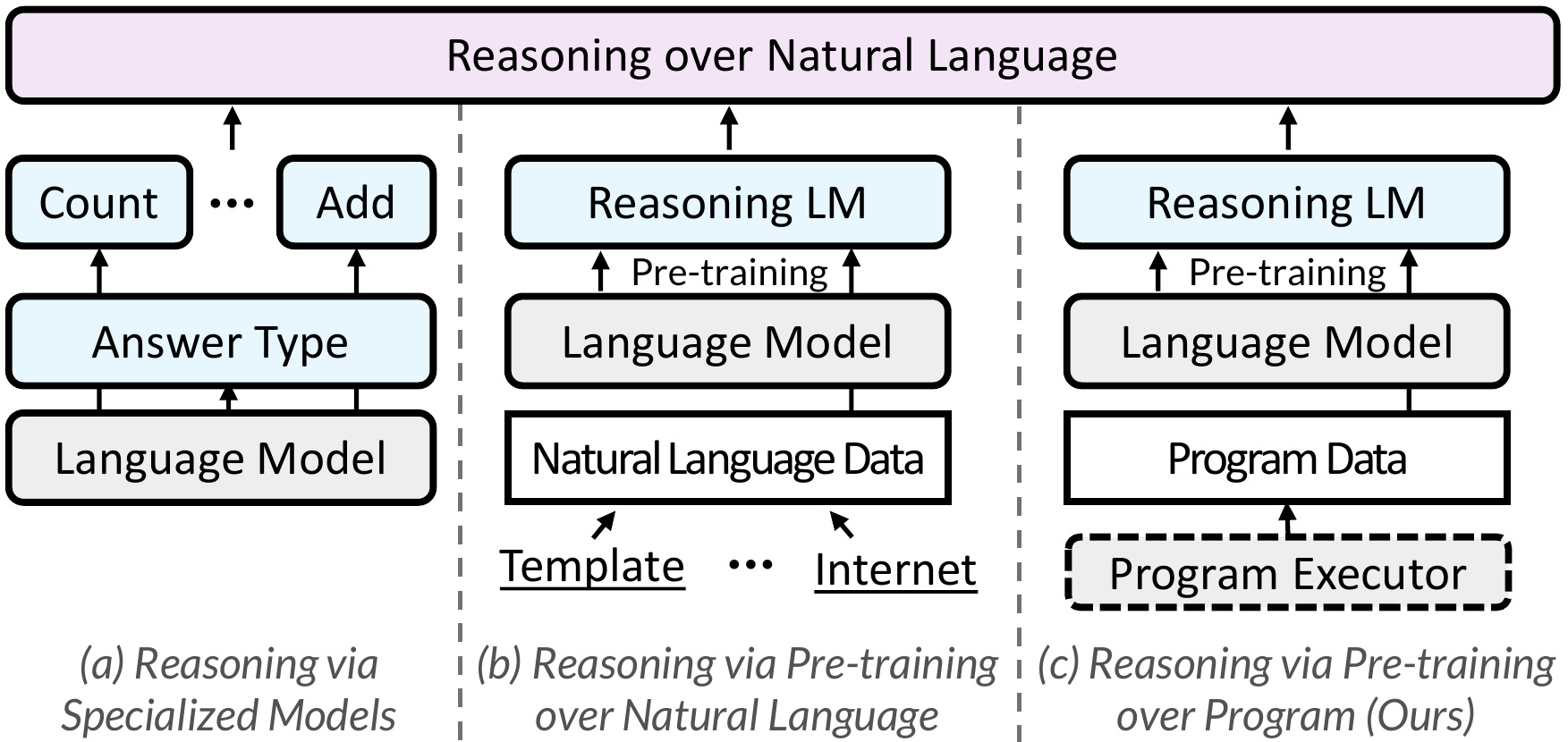}
    \caption{The illustration of different lines of reasoning, including \textbf{(a)} reasoning via specalized models, \textbf{(b)} reasoning via pre-training over natural language and \textbf{(c)} reasoning via pre-training over program (Ours).}
    \label{fig:related_work}
\end{figure*}

Since we focus on reasoning over natural language, our work is closely related to previous work on \textit{reasoning skills} in NL tasks.
Regarding methods to inject reasoning skills into LMs, our method is related to two lines of work contributing to the topic: \textit{specialized models} and \textit{pre-training}.
Last, our work is also related to \textit{program execution} since we employ program executors during pre-training.

\paragraph{Reasoning Skills}
The literature focuses on reasoning skills, including numerical reasoning~\citep{dua-etal-2019-drop,li-etal-2022-learning}, multi-hop reasoning~\citep{yang-etal-2018-hotpotqa}, reasoning in hybrid context~\citep{chen-etal-2020-hybridqa,zhu-etal-2021-tat} and logical reasoning~\citep{ijcai2020-501,yu2020reclor}.
Our work concentrates on improving the above reasoning skills, leaving the other reasoning abilities such as commonsense reasoning~\citep{zellers-etal-2018-swag,talmor-etal-2019-commonsenseqa,Bhagavatula2020Abductive} for future work.

\paragraph{Reasoning via Specialized Models}
Early works typically design specialized models and augment them into LMs for different types of questions~\citep{dua-etal-2019-drop,andor-etal-2019-giving,hu-etal-2019-multi,ding-etal-2019-cognitive}.
Taking \citet{hu-etal-2019-multi} as an example, they first predicted the answer type of a given question (e.g., ``how many''), and then adopted the corresponding module (e.g., count module) to predict the answer.
Although these methods work well on a specific dataset, it is challenging for them to scale to complex reasoning scenarios~\citep{Chen2020Neural}.
Differently, our work follows the line of reasoning via pre-training, which enjoys better scalability.

\paragraph{Reasoning via Pre-training}
This line of work focuses on the continued pre-training of LMs using large-scale data which involves reasoning.
The pre-training data are generally NL text, which are either crawled from Web with distant supervision~\citep{deng2021reasonbert}, generated by a model-based generator~\citep{asai-hajishirzi-2020-logic}, or synthesized via human-designed templates~\citep{geva-etal-2020-injecting,DBLP:journals/corr/abs-2107-07261,campagna-etal-2020-zero,wang2021logic}.
However, large-scale high-quality textual data involving reasoning is difficult to collect~\citep{deng2021reasonbert}.
Meanwhile, as the complexity of desired reasoning operations increases, synthesizing high-quality (e.g., fluent) NL sentences becomes more challenging.
Different from the above pre-training methods relying on NL data, our pre-training is performed on programs.
These programs can be synthesized at any scale with high quality, and thus are much easier to collect.

\paragraph{Reasoning in Giant Language Models}

Recent works demonstrate that with proper prompting (e.g., chain-of-thoughts prompting), giant language models (e.g., GPT-3) can perform well on reasoning tasks~\citep{wei2022chain,kojima2022large,li2022advance}.
For example, \citet{wei2022chain} find that giant language models have the ability to perform complex reasoning step by step with few-shot examples.
Although these prompting strategies do not need further fine-tuning, the basic assumptions of them and \ours are similar, i.e., it is difficult to obtain large amounts of clean sentences involving complex reasoning.
However, these prompting strategies do not work well for non-giant language models, while \ours is simultaneously applicable to language models ranging from millions (e.g., BART) to billions (e.g., T5-11B).
It is also interesting to investigate how these prompting strategies and \ours can be combined.

\paragraph{Program Execution}

We present a framework to leverage program executors to train LMs, and thus our work is close to recent work on learning a neural program executor.
In this line, the most related work to ours is \citet{liu2021tapex}, which revealed the possibility of SQL execution on helping table pre-training.
Different from them mainly focusing on table-related tasks, 
we present a generalized approach to include Math, Logic, and SQL, as well as their applications on many different natural language downstream tasks.
Other related studies include learning program executors on visual question answering \cite{andreas2016neural}, reading comprehension \cite{DBLP:journals/corr/abs-1912-04971, DBLP:journals/corr/abs-2009-00751}, knowledge base question answering~\cite{Ren2021LEGOLE} and 3D rendering \cite{tian2019learning}.
These works mainly focus on learning a neural network to represent the program executor, while ours focuses on transferring the knowledge of program executor to downstream tasks via pre-training.
Other lines of research leverage program execution in inference as a reliable sanity guarantee for generated programs by pruning non-executable candidates~\citep{Wang2018RobustTG,chen2018execution,chen-etal-2021-retrack,odena2020bustle, DBLP:conf/nips/EllisNPSTS19, chen2018execution, sun2018neural, DBLP:journals/corr/abs-1809-04682}.

\section{Reasoning Like Program Executors}\label{sec:reasoning}

Reasoning is the process where deduction and induction are sensibly applied to draw conclusions from premises or facts \citep{Scriven1976-SCRR-20}.
As a supreme feature of intelligence, humans apply reasoning across modalities.
Taking numerical reasoning as an example, humans can tell how many chocolates are consumed from a math word problem description, or from a real-world event where a mother gets off work and finds the choco-can empty, aside standing their guilty-looking kids with brownish stains on their faces.
Through detachment of information from their superficial modality and symbolic abstraction, humans manage to unify input formats and condense their numerical reasoning knowledge into one executable symbolic system -- This is the origin of an arithmetic program executor.
If a model can master these reasoning skills by imitating program executors, we believe in the possibility of transferring those reasoning skills to different modalities. 
In our case, we expect language models to transfer reasoning to NL-related tasks.
Given this motivation, we discuss the fundamental components of \ours in the rest of this section and present its instantiations later.

\paragraph{Program}\label{sec:para:program} refers to a finite sequence of symbols that can be understood and executed by machines.
For example, a program can be a logical form (e.g., Prolog), a piece of code (e.g., Python), or a math expression.
Compared with NL sentences, programs are more formal.
Each well-established program follows a specific set of grammar rules and can thus be synthesized systematically.
The generalizability of \ours framework is free from assumption and derived from the set of grammar rules on which a program follows.
In \ours, as long as a program returns meaningful output to reflect its computational procedure, it is an acceptable program.

\paragraph{Program Context} is the environment in which a program is running, which holds numerous variables accessible to the program. These variables serve as pivot points that anchor the program context with the program. In the same sense, the question and the passage in reading comprehension hold a similar relationship.
This suggests a natural analogy between the program-to-program context and the sentence-to-natural context in~\figref{overview}.

\paragraph{Program Executor} is a black-box software that can execute a given program within the program context.
An example could be the Python interpreter that executes each line of code, with its specific input data structures as program context.
For \ours, program executors play the role of teachers to educate students (i.e., LMs) on reasoning knowledge they contain.
\ours expects program executors to deterministically execute an input program with respect to a specific program context.

\begin{figure}[t]
    \centering
    \includegraphics[width=0.98\columnwidth]{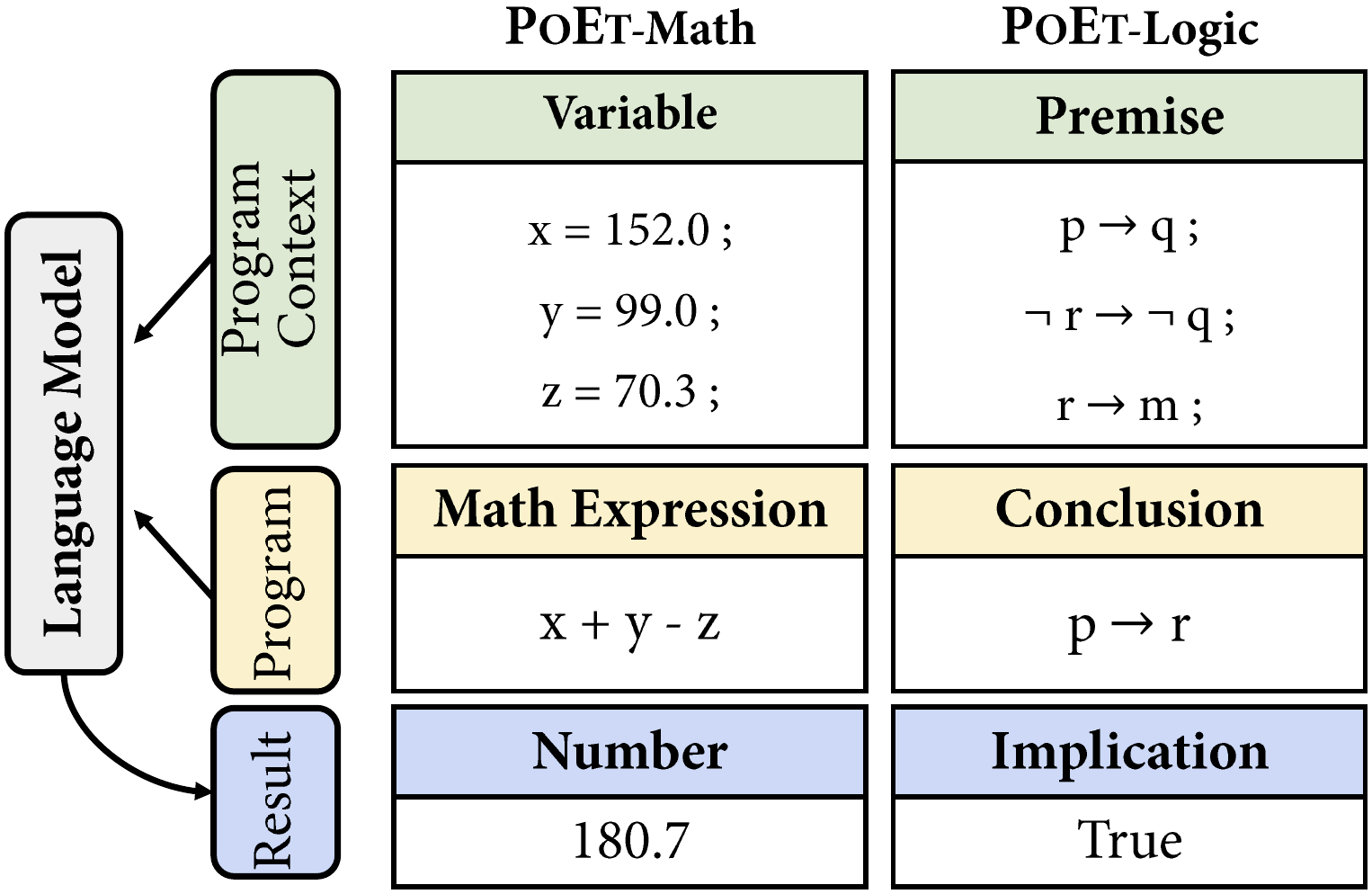}
    \caption{The illustration of \ours-Math and \ours-Logic. During pre-training, the concatenation of \textit{program} and \textit{program context} are fed into \textit{language model} and the model is expected to output \textit{result}.}
    \label{fig:program_context}
\end{figure}

\paragraph{Execution Result}
is obtained from the program executor, given a program and program context as input.
It is much analogous to the answer part in NL downstream tasks.
The execution result is the primary observable data reflecting the intermediate reasoning process and serves as the supervision provided by the program executor.

\section{\ours with Singleton Executors}\label{sec:application}

We first instantiate \ours with two singleton (i.e., a single type of reasoning capability) executors and then move on to \ours with integrated executors.

\subsection{Learning from Math Calculators}

The \ours-Math (Left in~\figref{program_context}) aims at injecting numerical reasoning skills into LMs via learning from math calculators.
Specifically, \ours-Math is designed to boost the basic arithmetic skills (i.e., addition and subtraction) of LMs on downstream tasks.
This arithmetic skill aligns with requirements to answer questions centered on addition\,/\, subtraction between two numbers, such as ``What is the difference in casualty numbers between Bavarian and Austrian?''.

\paragraph{Pre-training Task} Given several floating-point variables as the program context and a math expression only involving addition/ subtraction as the program, the pre-training task of \ours-Math is to \textit{calculate the math expression}.
Taking the leftmost example from ~\figref{program_context}, receiving the concatenation of the program and the program context as the input, \ours-Math is trained to output the number \texttt{\small 180.7}.
Considering the output can be an arbitrary number, the encoder-decoder model~\citep{lewis-etal-2020-bart} is more suitable for this pre-training task. 

\paragraph{Pre-training Corpus} Each example in the corpus contains a math expression containing up to $2$ operators and $3$ variables, and a program context that contains at most $30$ floating-point variables\,\footnote{More discussion can be found in Appendix~\secref{append_var}.}.
The mathematical addition and subtraction operators are denoted by \texttt{+} and \texttt{-}, respectively.
The values of variables vary from $0.0$ to $1000.0$.
By random generation, we synthesize $4$ million examples as the pre-training corpus for \ours-Math.

\subsection{Learning from Logic Solvers}

The \ours-Logic (Mid in~\figref{program_context}) aims at injecting logical reasoning (e.g., necessary conditional reasoning) skills into LMs via learning from logic solvers.
For example, taking the facts ``Only if the government reinforces basic education can we improve our nation's education to a new stage. In order to stand out among other nations, we need to have a strong educational enterprise.'' as premises, \ours-Logic is intended to help LMs identify whether the conclusion ``In order to stand out among nations, we should reinforce basic education'' is necessarily implied.

\begin{figure}[t]
    \centering
    \begin{tikzpicture}
    \scriptsize{
    \begin{axis}[
      ymajorgrids,
      xmajorgrids,
      grid style=dashed,
      xbar,
      height=.24\textwidth,
      width=.45\textwidth,
      bar width=1em,
      enlarge y limits=0.2,
      symbolic y coords={{POET-Logic},{RoBERTa-Large},{POET-Math},{BART-Large}},
      ytick distance=1,
      y tick style={opacity=0},
      bar shift=0pt,
      enlarge x limits=0.3,xticklabel style={/pgf/number format/fixed,/pgf/number format/fixed zerofill,/pgf/number format/precision=1},
      nodes near coords,
      nodes near coords align={horizontal}]
      \addplot[fill=blue!30, draw=blue] coordinates {(66.2,{BART-Large}) (75.2,{POET-Math})};
      \addplot[fill=red!30, draw=red] coordinates {(36.7,{RoBERTa-Large}) (38.9,{POET-Logic})};
    \end{axis}
  }
    \end{tikzpicture}
    \caption{Fine-tuning EM performance $[\%]$ of different models on DROP ({\color{blue!90}blue}) and LogiQA ({\color{red!90}red}).}
    \label{fig:pre}
  \end{figure}
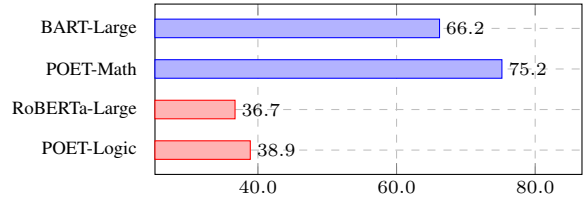

\paragraph{Pre-training Task} Given a few first-order logic premise statements as the program context and one conclusion statement as the program, the pre-training task of \ours-Logic is to identify \textit{if the program is necessarily implied from the program context}.
The execution result, i.e., the implication relationship between the program and the program context, is either \texttt{\small True} or \texttt{\small False}.
Since the output is binary, an encoder-only model \citep{DBLP:journals/corr/abs-1907-11692} is sufficient to perform this pre-training task.

\paragraph{Pre-training Corpus} Each example in the corpus contains several premise statements and a conclusion statement.
Initially, the statement collection for each example is empty.
To produce it, we first allocate $5$ Boolean variables (e.g., $p$ and $q$ in~\figref{program_context}) and randomly sample at most $8$ pairs from their pairwise combinations.
For each sampled pair $(p,q)$, we randomly select a statement from the set $\{p \rightarrow q,p \rightarrow {\neg}\,q,{\neg}\,p \rightarrow {\neg}\,q,{\neg}\,p \rightarrow q\}$ and add it to the collection.
Once the statement collection is prepared, we randomly select a statement as the conclusion statement (i.e., program) and the rest as the premise statements (i.e., program context).
Last, we employ Z3~\citep{de2008z3}, the well-known satisfiability modulo theory solver, as our program executor to obtain the implied result.
Finally, we synthesize $1$ million examples as the pre-training corpus for \ours-Logic, and nearly $16$\% examples\footnote{To clarify, 16\% is not a specific-purpose design but a statistical result.} correspond to \texttt{\small True}.

\subsection{Preliminary Observation}

We perform experiments on DROP and LogiQA to verify if our method improves the reasoning capability required by the dataset.
As observed in~\figref{pre}, \ours-Math boosts the numerical reasoning ability of BART, bringing in $9.0\%$ EM gain on DROP.
Meanwhile, \ours-Logic improves the logical reasoning skills of RoBERTa, resulting in a $2.2\%$ EM improvement on LogiQA.
These significant improvements support our main claim that reasoning knowledge of program executors can be transferred to NL scenarios via pre-training.

\begin{table}[t]
    \small
    \centering
    \begin{tabular}{cp{5.2cm}}
    \toprule
    \textbf{Type} & \textbf{Example SQL Program}  \\
    \midrule
    Arithmetic & \texttt{\scriptsize SELECT [COL]$_1$} - \texttt{\scriptsize [COL]$_2$} \\ \dline
    Superlative & \texttt{\scriptsize SELECT MAX([COL]$_1$)} \\ \dline
    Comparative &  \texttt{\scriptsize SELECT [COL]$_1$ WHERE [COL]$_2$ > [VAL]$_2$}\\ \dline
    Aggregation & \texttt{\scriptsize SELECT COUNT([COL]$_1$)} \\
    \dline
    Union & \texttt{\scriptsize SELECT [COL]$_1$ WHERE [COL]$_2$ = [VAL]$_2$ OR [COL]$_3$ = [VAL]$_3$} \\ \dline
    Nested & \texttt{\scriptsize SELECT [COL]$_1$ WHERE [COL]$_2$ IN ( SELECT [COL]$_2$ WHERE [COL]$_3$ = [VAL]$_3$)} \\
    \bottomrule
    \end{tabular}
    \caption{The six typical SQL programs that require reasoning. Listed are the type and the example SQL programs. \texttt{[COL]} and \texttt{[VAL]} represent the table column and the table cell value, respectively.}
    \label{tab:sql_program_type}
\end{table}

\section{\ours with Integrated Executors}

\ours-Math and \ours-Logic each focus on one specific reasoning skill, making the pre-training task heavily dependent on the downstream task.
Different from them, \ours-SQL is proposed to allow LMs to master different reasoning skills simultaneously.
In our implementation, \ours-SQL is pre-trained with an integrated SQL executor, since we believe that SQL queries are complex enough to encompass a wide variety of computational procedures (\tabref{sql_program_type}).

\paragraph{Pre-training Task}

\begin{figure}[t]
    \centering
    \includegraphics[width=0.98\columnwidth]{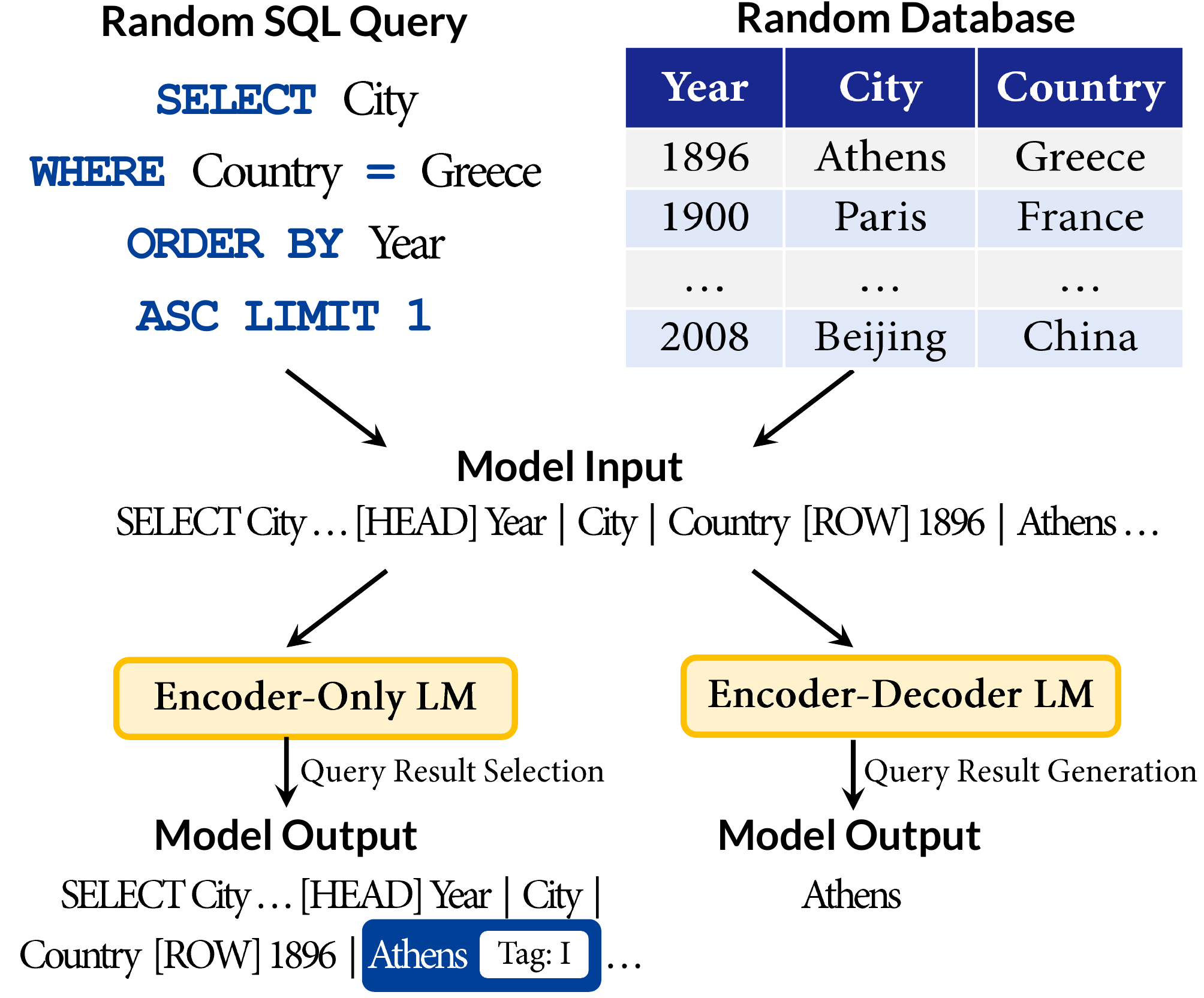}
    \caption{The illustration of \ourssql pre-training tasks: query result selection for encoder-only and query result generation for encoder-decoder LMs.}
    \label{fig:sql_overview}
\end{figure}

Given a SQL query as the program and a database as the program context, the pre-training task of \ours-SQL is to mimic the \textit{query result generation}.
As shown on the right side of~\figref{sql_overview}, given the concatenation of the program and the program context, the model is pre-trained to output the query result.
Since the encoder-decoder LMs can generate arbitrary tokens, they are well suited for the task.
On the other hand, encoder-only LMs have insufficient expressiveness to produce out-of-context query results.
To allow them to benefit from the SQL execution, we tailor the task into a \textit{query result selection} task for encoder-only LMs, which only utilizes query results that can be found in the database.
Specifically, the task requires encoder-only LMs to perform an \texttt{IO} sequence tagging process to find the query results in the database, as shown on the left side of~\figref{sql_overview}.
Note that the tag \texttt{I} is for tokens in the query results (e.g., Athens), while \texttt{O} is for other tokens.

\begin{table*}[t]
\small
\centering
\scalebox{0.86}{
\begin{tabular}{lllllllll}
    \toprule
      \multirow{2}{*}{\textbf{Models}}  & \multicolumn{2}{c}{\textbf{DROP}$^\heartsuit$} & \multicolumn{2}{c}{\textbf{HotpotQA}$^\heartsuit$} & \multicolumn{2}{c}{\textbf{TAT-QA}$^\heartsuit$} & \textbf{SVAMP} & \textbf{EQUATE} \\
        \cmidrule(lr){2-3}
        \cmidrule(lr){4-5}
        \cmidrule(lr){6-7}
        \cmidrule(lr){8-8}
        \cmidrule(lr){9-9}
        & EM & F1 & EM & F1 & EM & F1 & EM & EM \\
        \toprule
        BART-Large & $66.2$ & $69.2$ & $65.6$ & $78.9$ & $38.8$ & $46.7$ & $12.4$ & $62.6$\\
        \oursbart & $77.7$ (+$11.5$) & $80.6$ (+$11.4$) & $66.5$ (+$0.9$) & $79.7$ (+$0.8$) & $41.5$ (+$2.7$) & $49.6$ (+$2.9$) & $33.5$ (+$21.1$) & $66.5$ (+$3.9$)\\
        \midrule
        RoBERTa-Large & $\underline{78.1}$  & $\underline{85.3}$ & $67.6$ & $81.1$ & $55.2$ & $62.7$ & -- & $64.2$ \\
        \oursroberta & $\underline{79.8}$ (+$1.7$) & $\underline{87.4}$ (+$2.1$) & $68.7$ (+$1.1$) & $81.6$ (+$0.5$) & $59.1$ (+$3.9$) & $65.9$ (+$3.2$)  & -- & $67.5$ (+$3.3$)\\
        \bottomrule
\end{tabular}}
\caption{The main experimental results of different backbone models on test sets and dev sets ($\heartsuit$) of datasets with or without our \ourssql. The results of \ours are significantly better than the vanilla LMs ($p<0.05$). Note the performance of RoBERTa and \oursroberta are reported on the \underline{subset} of DROP where the answer is span(s).}
\label{tab:backbone-improve}
\end{table*}

\begin{table*}[t]
    \small
    \begin{subtable}[h]{0.48\textwidth}
    \centering
    \scalebox{0.9}{
        \begin{tabular}{lccc}
            \toprule
         {\textbf{Dataset}} &  {\textbf{Models}} & \textbf{EM} & \textbf{\fone{}} \\
        \toprule
        \multirow{12}{*}{DROP$^\heartsuit$}
        & \multicolumn{3}{c}{\textit{Specialized Models}} \\
        & NumNet~\citep{ran-etal-2019-numnet} & $64.9$ & $68.3$ \\
        & MTMSN~\citep{hu-etal-2019-multi} & $76.7$ & $80.5$ \\
        & NeRd~\citep{Chen2020Neural} & $78.6$ & $81.9$ \\
        & NumNet+~\citep{ran-etal-2019-numnet} & $81.1$ & $84.4$ \\
        & QDGAT~\cite{chen-etal-2020-question} & $\mathbf{84.1}$ & $\mathbf{87.1}$ \\
        & \multicolumn{3}{c}{\textit{Language Models}} \\
        & T5~\citep{DBLP:journals/corr/abs-2107-07261} & $61.8$ & $64.6$ \\
        & GenBERT~\citep{geva-etal-2020-injecting} & $68.8$ & $72.3$ \\
        & PReasM~\citep{DBLP:journals/corr/abs-2107-07261} & $69.4$ & $72.3$ \\
        & \oursbart & ${77.7}$ & ${80.6}$ \\
        & {\textsc{PoEt}-Math+SQL$_{\rm {\color{Magenta} {\tt{BART}}}}$} & $\mathbf{78.0}$ & $\mathbf{80.9}$ \\
        \midrule
        \multirow{4}{*}{TAT-QA$^\heartsuit$}
        & \tapas~\citep{herzig-etal-2020-tapas} & $18.9$ & $26.5$ \\
        & NumNet+~\citep{ran-etal-2019-numnet} & $38.1$ & $48.3$ \\
        & \tagop~\citep{zhu-etal-2021-tat} & ${55.2}$ & ${62.7}$ \\
        & \tagop + \oursroberta & $\mathbf{59.1}$ & $\mathbf{65.9}$ \\
        \bottomrule
        \end{tabular}
        }
    \end{subtable}
    \hfill
    \begin{subtable}[h]{0.52\textwidth}
    \scalebox{0.9}{
        \begin{tabular}{lccc}
        \toprule
         {\textbf{Dataset}} &  {\textbf{Models}} & \textbf{EM} & \textbf{\fone{}} \\
        \toprule
        \multirow{11}{*}{HotpotQA$^\heartsuit$}
        &  \multicolumn{3}{c}{\textit{Specialized Models}} \\
        & DFGN~\citep{qiu-etal-2019-dynamically} & $55.7$ & $69.3$ \\
        & SAE~\citep{DBLP:conf/aaai/TuHW0HZ20} & $67.7$ & $80.8$ \\
        & C2F Reader~\citep{shao-etal-2020-graph} & $68.0$ & $81.2$ \\
        & HGN~\citep{fang-etal-2020-hierarchical} & $\mathbf{69.2}$ & $\mathbf{82.2}$ \\
        & \multicolumn{3}{c}{\textit{Language Models}} \\
        & BERT~\citep{devlin-etal-2019-bert} & $59.1$ & $73.4$ \\
        & ReasonBERT~\citep{deng2021reasonbert} & $64.8$ & $79.2$ \\
        & \oursbart & $66.5$ & $79.7$ \\
        & SpanBERT~\citep{joshi-etal-2020-spanbert} &  $67.4$ & $81.2$ \\
        & \oursroberta & $\mathbf{68.7}$ & $\mathbf{81.6}$ \\
        \midrule
        \multirow{5}{*}{EQUATE}
        & BERT~\citep{devlin-etal-2019-bert} & $51.8$ & -- \\
        & GPT~\citep{radford2019language} & $55.8$ & -- \\ 
        & QREAS~\citep{ravichander-etal-2019-equate} & $60.7$ & -- \\
        & \oursbart & $66.5$ & -- \\
        & \oursroberta & $\mathbf{67.5}$ & -- \\
        \bottomrule
        \end{tabular}
    }
    \end{subtable}
    \caption{The comparison of our models with baselines on test sets and dev sets ($\heartsuit$) of different datasets.
    LMs used by all baselines are in Large size, except for ReasonBERT. Bold numbers indicate the best results.}
    \label{tab:overall-metric}
\end{table*}

\paragraph{Pre-training Corpus}

Each example in the corpus contains a SQL query, a database, and a query result.
Notably, following \citet{liu2021tapex}, each database is flattened into a sequence when it is fed into LMs.
Meanwhile, to avoid databases being too large to fit into memory, we randomly drop the rows of large databases until their flattened sequences contain less than $450$ tokens.
For the query result generation task, we follow the same corpus construction strategy as described in \citet{liu2021tapex}.
Concretely, by instantiating SQL templates from \textsc{Squall}~\cite{shi-etal-2020-potential} over databases provided by \textsc{WikiSQL}~\cite{DBLP:journals/corr/abs-1709-00103}, $5$ million examples are synthesized for pre-training.
For the query result selection task, the pre-training corpus is constructed in a similar way as above, except that only the examples whose query results are suitable for encoder-only are retained.
This filtering results in a corpus containing nearly $2$ million examples.

\section{Experiments and Analysis}\label{sec:expr}

To verify the effectiveness of \ourssql on boosting the reasoning capabilities of LMs, we first apply our method on several backbone models, including encoder-only models and encoder-decoder models.
Then we conduct experiments on five typical reasoning benchmark datasets and compare \ourssql with previous methods.
Last, we perform a detailed model analysis to provide more insights.

\paragraph{Dataset Setup} 

We perform experiments on different datasets including DROP, HotpotQA, TAT-QA, and EQUATE.
\tabref{discrete_reasoning} shows examples of these datasets and their corresponding reasoning types.
Furthermore, SVAMP~\citep{patel-etal-2021-nlp}, the challenging diagnostic dataset for probing \textit{numerical reasoning}, is employed in our experiments to test the generalization capability of our fine-tuned models on DROP.
We evalute models on their addition and subtraction subsets.
More details about datasets can be found in Appendix~\secref{append_dataset}.

\paragraph{Backbone Model}
RoBERTa~\cite{DBLP:journals/corr/abs-1907-11692} is elected as the backbone in encoder-only LMs, while BART~\cite{lewis-etal-2020-bart} is chosen as the backbone in encoder-decoder LMs.
Afterward, We mark the RoBERTa model and the BART model trained under \ours as \oursroberta and \oursbart, respectively.
For \oursbart, we treat all datasets as generative tasks and fine-tune models to generate answers.
As for \oursroberta, the fine-tuning strategies on different datasets are slightly different, and more implementation details can be found in Appendix~\secref{append_imp}.
Notably, on all datasets, our models are evaluated with official evaluation metrics EM and F1.

\subsection{Experimental Results}\label{sec:method_compare}

\paragraph{Comparing to Vanilla LMs}

\tabref{backbone-improve} presents an apple-to-apple performance comparison between \ourssql models and their associated vanilla LMs.
Across all instances, we observe significant performance increment on downstream NL reasoning tasks.
Specifically, \ourssql equips popular encoder-only and encoder-decoder models with an integrated package of reasoning skills, effectively improving their performance on five benchmark datasets.
As a highlighted example, \oursbart obtains $11.5\%$ (DROP) and $21.1\%$ (SVAMP) improvements on EM, compared with the vanilla BART.
Since \ours pre-training is carried purely on program context, whereas all downstream tasks are on natural context, our hypothesis that reasoning capability is transferable from program executors to NL scenarios gets verified.

\paragraph{Comparing to Previous Methods}

\tabref{overall-metric} lists all experimental results of baselines and our models on different datasets.
As seen, our model generally achieves highly competitive results on different reasoning skills, showing its strong performance.
Compared with other reasoning-enhanced LMs, \oursbart surpasses them by a large margin, demonstrating the effectiveness of our proposed program execution pre-training.
For example, compared with PReasM initialized from T5-Large, \oursbart initialized from BART-Large exceeds it by $8.3\%$.
Furthermore, \ours that learns from a mix of program executors (i.e., {\textsc{PoEt}-Math+SQL$_{\rm {\color{Magenta} {\tt{BART}}}}$}) achieves a slightly better performance than the single prgoram executor.

\subsection{Pre-training Analysis}

We show part of the analysis results below due to the limited space, and more analysis can be found in Appendix~\secref{append_var} and~\secref{fine_analysis}.

\definecolor{upurple}{RGB}{155,89,182}
\definecolor{ublue}{RGB}{52,152,219}
\definecolor{ured}{RGB}{231,76,60}
\definecolor{udark}{RGB}{77,153,77}
\definecolor{ugreen}{RGB}{46,204,113}

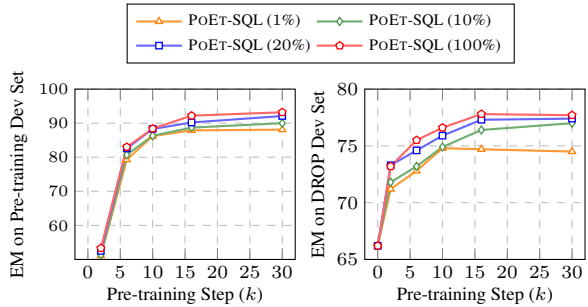
\begin{figure}[t!]
    \centering
        \begin{tikzpicture}
        \scriptsize
        \begin{axis}[
        width=.28\textwidth, height=.24\textwidth ,
        ymin=50, ymax=100,
        xlabel style={yshift=0ex,},
        xmin=-2, xmax=32,
        xtick={0,5,10,...,30},
        ytick={60,70,80,90,100},
        y tick style={opacity=0},
        xlabel={{Pre-training Step ($k$)}},
        xlabel style={align=center,font=\scriptsize,yshift=1em},
        grid style=dashed,
        ylabel={EM on Pre-training Dev Set},
        xlabel style={align=center,font=\scriptsize},
        y tick label style={font=\tiny},
        xlabel style={align=center,font=\scriptsize},
        ylabel style={font=\scriptsize,yshift=-2em},
        ymajorgrids=true,
        xmajorgrids=true,
        tick align=inside]
        \addplot[orange!80,mark=triangle*,,mark size=1.5pt,thick,mark options={fill=white,draw=orange,line width=0.5pt}]
            coordinates {
              (2, 51.1)
              (6, 79.3)
              (10, 86.2)
              (16, 87.9)
              (30, 88.1)
            };

        \addplot[
            udark!80,mark=diamond*,mark size=1.5pt,thick,mark options={fill=white,draw=udark,line width=0.5pt}]
            coordinates {
            (2, 51.4)
            (6, 80.7)
            (10, 86.3)
            (16, 88.7)
            (30, 90.0)
            };

        \addplot[
            blue!60,mark=square*,mark size=1.2pt,thick,mark options={fill=white,draw=blue,line width=0.5pt}
            ]
            coordinates {
            (2, 52.5)
            (6, 82.6)
            (10, 88.3)
            (16, 90.2)
            (30, 92.1)
            };

        \addplot[
            red!60,mark=pentagon*,mark size=1.5pt,thick,mark options={fill=white,draw=red,line width=0.5pt}
            ]
            coordinates {
            (2, 53.3)
            (6, 83.0)
            (10, 88.4)
            (16, 92.2)
            (30, 93.2)
            };
        \end{axis}
        \vspace{6cm}
        \begin{axis}[
        at={(15.5em,0)},
        width=.28\textwidth, 
        height=.24\textwidth,
           ymin=65, ymax=80,
            xlabel style={yshift=0ex,},
            xmin=-2, xmax=32,
            ymajorgrids,
            xmajorgrids,
            tick align=inside,
            grid style=dashed,
            y tick style={opacity=0},
            xtick={0,5,10,...,30},
            ytick={65, 70, 75, 80},
            yticklabels={$65$, $70$, $75$, $80$},
            xlabel={{Pre-training Step ($k$)}},
            ylabel={{EM on DROP Dev Set}},
            xlabel style={align=center,font=\scriptsize,yshift=1em},
            ylabel style={font=\scriptsize,yshift=-2.5em},
            legend style={at={(-0.2,1.15)},anchor=south},
            legend columns=2,
            legend cell align={left},
            clip=false
        ]
         
        \addplot[orange!80,mark=triangle*,,mark size=1.5pt,thick,mark options={fill=white,draw=orange,line width=0.5pt}]
            coordinates {
              (0, 66.2)
              (2, 71.2)
              (6, 72.8)
              (10, 74.8)
              (16, 74.7)
              (30, 74.5)
            };
        \addlegendentry{\tiny{\ourssql($1$\%)}}
            
        \addplot[
            udark!80,mark=diamond*,mark size=1.5pt,thick,mark options={fill=white,draw=udark,line width=0.5pt}]
            coordinates {
            (0, 66.2)
            (2, 71.8)
            (6, 73.2)
            (10, 74.9)
            (16, 76.4)
            (30, 77.0)
            };
            \addlegendentry{\tiny{\ourssql($10$\%)}}
            
        \addplot[
            blue!60,mark=square*,mark size=1.2pt,thick,mark options={fill=white,draw=blue,line width=0.5pt}
            ]
            coordinates {
            (0, 66.2)
            (2, 73.3)
            (6, 74.6)
            (10, 75.9)
            (16, 77.3)
            (30, 77.4)
            };
            \addlegendentry{\tiny{\ourssql($20$\%)}}
        
        \addplot[
            red!60,mark=pentagon*,mark size=1.5pt,thick,mark options={fill=white,draw=red,line width=0.5pt}
            ]
            coordinates {
            (0, 66.2)
            (2, 73.2)
            (6, 75.5)
            (10, 76.6)
            (16, 77.8)
            (30, 77.7)
            };
            \addlegendentry{\tiny{\ourssql($100$\%)}}
        \end{axis}
        \end{tikzpicture}
    \caption{The EM performance $[\%]$ on the pre-training dev set (\textbf{Left}) and the downstream DROP dev set (\textbf{Right}) with different pre-training steps and scales. \ourssql ($x\%$) denotes the model trained with $x\%$ pre-training examples, while $100\%$ corresponds to the model trained with the whole pre-training corpus of \ourssql, which contains $5$ million examples.}
    \label{fig:scale_analysis}
\end{figure}

\paragraph{Necessity of Program Execution}

\ours hypothesizes that the acquisition of reasoning ability by models happens at the stage of mimicking program execution, rather than program language modeling.
To verify it, we ablate the program executor in \oursbart and carry out a SQL language modeling pre-training instead.
Practically, we mask each SQL query in the pre-training corpus of \ours-SQL using the strategy adopted in BART \citep{lewis-etal-2020-bart}, and pre-train BART to output the complete SQL query given the masked SQL query and the database.
The trivial performance variance corroborates the necessity of program execution.

\paragraph{Effect of the Pre-training Step and Scale}

\figref{scale_analysis} illustrates the pre-training and downstream performance with different pre-training steps and scales.
It can be seen that both pre-training and downstream performance gradually increase towards the asymptote with increasing pre-training steps, while extra pre-training data steadily accelerate the convergence rate.
Although a larger scale yields better performance on the pre-training dev set, $10\%$ ($500k$) data can already converge approximately to the same asymptote as the full data pre-training, showing the high data efficiency of \ours.
The highly plateaued curve also serves as sound evidence that execution pre-training is a data-efficient pre-training approach that converges quickly.

\pgfplotstableread[row sep=\\,col sep=&]{
    Dataset & wtq & sqa \\
    SQuAD   & 93.3 & 93.2 \\
    MNLI   & 90.8 & 89.6 \\
    QuoRef   & 84.9 & 84.7 \\
    }\mydata

\pgfplotsset{every axis/.append style={
                    label style={font=\large},
                    tick label style={font=\large} 
                    }}
\begin{figure}[t]
    \centering
    \begin{tikzpicture}[scale=0.6]
        \begin{axis}[
                ybar=5pt,
                bar width=.6cm,
                enlarge x limits=0.2,
                width=.5\textwidth,
                height=.3\textwidth,
                legend style={at={(1.35,0.5)},
                anchor=north,legend columns=1},
                symbolic x coords={SQuAD, MNLI, QuoRef},
                xtick=data,
                nodes near coords,
                x tick style={opacity=0},
                nodes near coords align={vertical},
                ymin=70,ymax=100,
                ylabel={Task Performance (\%)},
            ]
            \addplot table[x=Dataset,y=wtq]{\mydata};
            \addplot table[x=Dataset,y=sqa]{\mydata};
            \legend{RoBERTa-Large,\oursroberta}
        \end{axis}
    \end{tikzpicture}
    \caption{The performance comparison between RoBERTa-Large and \oursroberta on representative NLU tasks. On SQuAD and QuoRef, we report \fone{}, whereas on MNLI we report accuracy.}
    \label{fig:analysis_scale}
\end{figure}
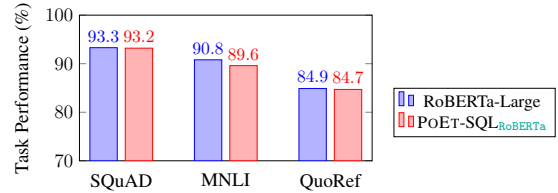

\section{Discussion and Open Questions}

In this section, we carry out comprehensive studies on \ours, summarize interesting open questions, and provide insights for future work.

\finding{Does \ours improve reasoning skills at the sacrifice of NL understanding abilities? \textbf{No}.}
\noindent
During pre-training, our models are exposed to artificial programs that are dramatically different from NL sentences, raising the concern that models may catastrophically forget their original NL understanding ability.
We explore this by comparing \oursroberta and the vanilla RoBERTa model on tasks focusing on NL understanding, including SQuAD, MNLI and QuoRef.
As can be observed in \figref{analysis_scale}, \oursroberta performs almost equally well as RoBERTa on two datasets (i.e., SQuAD and QuoRef), which suggests that \ours barely sacrifices the intrinsic NL understanding ability of LMs.
And the performance drop on MNLI (1.2\%) is also noteworthy, which may be alleviated by joint pre-training on language modeling and our proposed program execution.
More experimental details can be found in Appendix~\secref{nl_understnad}.

\begin{table}
\centering
\small
\scalebox{1.0}{
\begin{tabular}{lccc}
\toprule
\textbf{Settings} & EM & \fone{}  \\
\toprule
\oursbart & $77.7$ & $80.6$ \\
\multicolumn{3}{c}{\textit{Tuning Program}} \\
~~$\hookrightarrow$ \textit{w.} Nnatural program & $77.2$ & $79.9$ \\
~~$\hookrightarrow$ \textit{w.} Unnatural program & $76.9$ & $79.7$ \\
\multicolumn{3}{c}{\textit{Tuning Program Context}} \\
~~$\hookrightarrow$ \textit{w.} Natural program context & $76.5$ & $79.0$ \\
\bottomrule
\end{tabular}}
\caption{The EM and \fone{} of \oursbart on the DROP dev set with respect to different naturalness of program and program context.}
\label{tab:ablation}
\end{table}

\finding{Will \ours be affected by naturalness of program context or program? \textbf{No}.}
\noindent
An intuitive hypothesis is that the effectiveness of \ours should be positively associated with the naturalness of program and program context, due to closer learning objectives.
To test this hypothesis, we design two groups of experiments:
(i) Tuning the naturalness of program - we follow \citet{liu2021tapex} to translate SQL queries into NL sentences to make a more natural program, and replace SQL reserved keywords with low-frequency tokens to make a more unnatural program.
(ii) Tuning the naturalness of program context - we follow \citet{chen2019tabfact} to convert each database in \ourssql pre-training into a set of NL sentences.
Surprisingly, results in~\tabref{ablation} provide counter-evidence to the intuitive hypothesis, since tuning the naturalness of program or program context do not significantly impact \ours effectiveness.
For example, unnatural program only leads to a slight decrease in DROP EM from $77.7\%$ to $76.9\%$.
It also indictaes that the model learns certain abstract reasoning capabilities rather than lexical associations.

\begin{figure}[!t]
\centering
\begin{tikzpicture}
\scriptsize{
\begin{axis}[
at={(0,0)},
width=.28\textwidth, height=.22\textwidth ,
xtick={0,4,8,...,20},
x tick label style={font=\scriptsize},
ytick={1.0, 1.2, 1.4, 1.6, 1.8, 2.0},
grid style=dashed,
ylabel={Train\ \ Perplexity},
xlabel={{Training Step ($k$)}},
xlabel style={align=center,font=\scriptsize,yshift=1em},
ylabel style={font=\scriptsize,yshift=-2em},
y tick style={opacity=0},
y tick label style={font=\scriptsize},
ymajorgrids=true,
xmajorgrids=true,
tick align=inside,
legend pos=outer north east,
yticklabel style={/pgf/number format/precision=1,/pgf/number format/fixed zerofill},
legend style={yshift=-0.5em,xshift=-9.7em,legend cell align=left,legend plot pos=right},
xmin=-1,
xmax=22,
ymin=1.00,
ymax=2.00]
    \addplot[
        red!60,mark=pentagon*,mark size=1.2pt,thick,mark options={fill=white,draw=red,line width=0.5pt}
        ]
        coordinates {
        (2, 1.84)
        (4, 1.4)
        (6, 1.39)
        (8, 1.24)
        (10, 1.21)
        (12, 1.19)
        (14, 1.18)
        (16, 1.17)
        (18, 1.15)
        (20, 1.15)
        };
      \addplot[
        blue!60,mark=square*,mark size=1.2pt,thick,mark options={fill=white,draw=blue,line width=0.5pt}
        ]
        coordinates {
        (2, 1.61)
        (4, 1.34)
        (6, 1.28)
        (8, 1.22)
        (10, 1.18)
        (12, 1.16)
        (14, 1.15)
        (16, 1.14)
        (18, 1.13)
        (20, 1.13)
        };
        \legend{\tiny{BART},\tiny{BART+DROP}}
\end{axis}
}
\vspace{6cm}
\scriptsize{
\begin{axis}[
at={(15.5em,0)},
width=.28\textwidth, height=.22\textwidth ,
xtick={0,5,10,15,20},
x tick label style={font=\scriptsize},
ytick={1.1, 1.2, 1.3, 1.4, 1.5, 1.6, 1.7},
grid style=dashed,
xlabel={{Training Step ($k$)}},
ylabel={Dev\ \ Perplexity},
xlabel style={align=center,font=\scriptsize,yshift=1em},
ylabel style={font=\scriptsize,yshift=-2.5em},
y tick style={opacity=0},
y tick label style={font=\tiny},
ymajorgrids=true,
xmajorgrids=true,
tick align=inside,
legend pos=outer north east,
legend style={yshift=-0.5em,xshift=-9.7em,legend cell align=left,legend plot pos=right},
yticklabel style={/pgf/number format/precision=1,/pgf/number format/fixed zerofill},
xmin=-1,
xmax=22,
ymin=1.10,
ymax=1.70]

    \addplot[
        red!60,mark=pentagon*,mark size=1.2pt,thick,mark options={fill=white,draw=red,line width=0.5pt}
        ]
        coordinates {
        (2, 1.61)
        (4, 1.34)
        (6, 1.29)
        (8, 1.25)
        (10, 1.23)
        (12, 1.24)
        (14, 1.25)
        (16, 1.26)
        (18, 1.26)
        (20, 1.26)
        };
      \addplot[
        blue!60,mark=square*,mark size=1.2pt,thick,mark options={fill=white,draw=blue,line width=0.5pt}
        ]
        coordinates {
        (2, 1.41)
        (4, 1.29)
        (6, 1.26)
        (8, 1.23)
        (10, 1.22)
        (12, 1.21)
        (14, 1.22)
        (16, 1.23)
        (18, 1.24)
        (20, 1.24)
        };
        \legend{\tiny{BART},\tiny{BART+DROP}}
\end{axis}
}
\end{tikzpicture}
\caption{The train and dev perplexity of vanilla BART and BART pre-trained on DROP (BART+DROP) on the pre-training corpus of \ourssql.}
\label{fig:nl_transfer}
\end{figure}
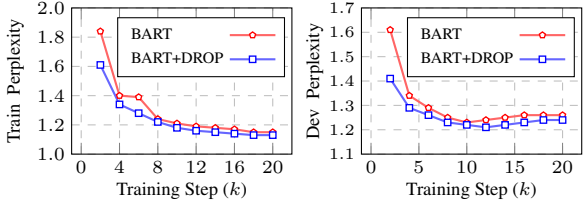

\finding{Does pre-training on NL reasoning benefit model learning on program execution? \textbf{Yes}.}
\noindent
If reasoning ability can be transferred from program execution to NL reasoning tasks in \ours, then the reversed process of \ours may also work, i.e., models pre-trained with NL reasoning would have better learnability on program execution.
To test this speculation, we compare the behavioral difference of vanilla BART and BART pre-trained on DROP in terms of learning SQL execution in~\figref{nl_transfer}.
There are two indicators that can be used to characterize the behavior of LMs on the SQL execution task: execution accuracy and perplexity, and the execution accuracy always goes higher when the perplexity goes lower.
Here the perplexity is presented because it is smoother compared to the execution accuracy, which is either 100\% or 0\%.
Parallel with our expectation, pre-training on DROP leads to observably lower perplexity for SQL execution learning on both the train and dev sets.
The bidirectional enhancement suggests some relative independence between reasoning mechanisms and their symbolic representations.

\finding{Can \ours boost reasoning abilities of giant pre-trained language models? \textbf{Yes}.}
\noindent
Recent work suggest that giant LMs excel at reasoning \cite{gpt3}, so we are curious if \ours is effective for them.
Following the same procedure as in~\secref{expr}, we apply \ourssql to T5-11B, one of the largest publicly available LMs.
As shown in~\tabref{tfive_improve}, albeit not as shining as in cases of smaller LMs, \ours still succeeds in boosting numerical reasoning abilities of giant LMs.

\begin{table}[bt]
\small
\centering
\scalebox{0.95}{
\begin{tabular}{llllll}
    \toprule
      \multirow{2}{*}{\textbf{Models}}  & \multicolumn{2}{c}{\textbf{DROP}$^\heartsuit$} & \textbf{SVAMP} \\
        \cmidrule(lr){2-3}
        \cmidrule(lr){4-4}
        & EM & F1 & EM \\
        \toprule
        T5-$11$B & $83.5$ & $85.9$ & $52.9$ \\
        \ourstfive & $85.2$ (+$1.7$) & $87.6$ (+$1.7$) & $57.4$ (+$4.5$) \\
        \bottomrule
\end{tabular}}
\caption{The experimental results of T5-11B and \ourstfive on test sets and dev sets ($\heartsuit$) of different datasets.}
\label{tab:tfive_improve}
\end{table}

\section{Conclusion \& Future Work}

We introduce \ours, a new pre-training paradigm for boosting reasoning capability of language models via imitating program executors.
Experimental results on six datasets demonstrate that \ours can significantly boost existing language models on several reasoning skills, including numerical, logical and multi-hop reasoning.
Our best language model under \ours can reach highly competitive performance with previous specialized models.
In the future, we hope our work could inspire more transference of reasoning knowledge from program executors to models.
And we will also investigate the causes of the reasoning transfer with more insightful experiments, since we still do not know how the reasoning transfer occurs.

\section*{Limitations}

The first limitation of our approach is that it has a relatively strong coupling between the reasoning skills learned in the pre-training task and the reasoning skills required for the downstream task.
In other words, \ours expects reasoning abilities of the program executor to overlap with the downstream reasoning requirements to make the execution learning transferable.
Such an expectation also applied fo \ourssql, although it allows LM to master different reasoning skills at the same time.
For example, when ablating all programs involving math operations from the pre-training corpus of \ourssql, it shows poor performance on DROP.
The second limitation is that \ours still employs instantiated program templates rather than probabilistic context-free grammars to synthesize programs.
The latter usually offers a more diverse range of programs that may contribute to the generalization of the pre-trained language models, but are often more complex.

\section*{Acknowledgement}

We would like to thank Frank F. Xu, Zhengbao Jiang, Bill Yuchen Lin, Shuyan Zhou, Zhiruo Wang, Libo Qin, Jize Jiang, and Jonathan Livengood for fruitful discussion.
We also thank all the anonymous reviewers for their constructive feedback and insightful comments.

\bibliography{custom,anthology}
\bibliographystyle{acl_natbib}

\appendix

\section{Program Context Analysis}\label{sec:append_var}

\ours emphasizes the importance of program context for reasoning transferability, owing to the analogy between the program to program context and the sentence to natural context drawn in~\figref{overview}.
To investigate it, we explore the effect of different program context design choices on reasoning transferability by conducting experiments on well-designed \oursmath variants.

\subsection{The Necessairty of Program Context}

\begin{table}[tbp]
  \small
  \centering
  \scalebox{0.98}{
    \begin{tabular}{lcc}
    \toprule
    \textbf{Models} & \textbf{EM} & \textbf{F1} \\
    \midrule
     BART-Large  & $66.2$  & $69.2$  \\
     \midrule
     \oursbartmath \textit{without} program context & $67.4$ & $70.5$ \\ 
     \oursbartmath \textit{with} ~~$0$ irrelevant variable & $71.5$  & $74.5$  \\
     \oursbartmath \textit{with} $10$ irrelevant variables & $74.6$  & $77.5$  \\
     \oursbartmath \textit{with} $30$ irrelevant variables & $75.2$  & $78.1$  \\
    \bottomrule
    \end{tabular}%
  }
  \caption{The DROP performance with different numbers of irrelevant variables in \oursbartmath pre-training.}
  \label{tab:irr}%
\end{table}%

To verify the necessairty of program context, we experiment with \oursmath without program context, i.e. a variable-free \oursmath variant whose program context is empty.
Taking the example of \ours-Math in~\figref{program_context}, the program is transformed into \texttt{\small 152.0\,+\,99.0\,-\,70.3}.
The experimental results are shown in \tabref{irr}.
One can see that there is a dramatic performance drop in the variant compared to \oursmath, verifying the importance of program context.

\subsection{The Variables Design in Program Context}

In the pre-training task of \ours-Math, the program context is several floating-point variables.
These variables include necessary variables (i.e., variables required by the program) and irrelevant variables.
The irrelevant variables exist to make the program context closer to the natural context which generally contains irrelevant sentences.
For example, given the program \texttt{\small a + b} and the program context \texttt{\small a = 1; b = 2; c = 3; d = 4;}, variables \texttt{\small c} and \texttt{\small d} are what we refer to as irrelevant variables.
This is motivated by the fact that passages are usually full of irrelevant information regarding a specific question in NL downstream tasks.

In this section, we explore impacts on pre-training effectiveness brought by numbers of irrelevant variables.
Empirically, we experiment on pre-training with $0$, $10$, $30$ irrelevant variables.
The total length of $30$ irrelevant variables approaches the maximum input length of pre-trained LMs, and thus we do not try more settings.
The experimental results are shown in \tabref{irr}.
As observed, (i) models can still learn numerical reasoning during pre-training where the program context is free from irrelevant variables, though less effective. (ii) the setting of $30$ irrelevant variables brings BART-Large more performance improvement than the setting of $10$ irrelevant variables.
Considering there are plenty of lengthy passages in the DROP dataset, we therefore hypothesize that the noise level brought by irrelevant variables in the program context during pre-training should be made closer with the counterpart in the natural context during fine-tuning.

\section{Experimental Setup}\label{sec:append_dataset}

\subsection{Dataset Setup}

\tabref{dataset-data} presents some statistics about our experimental datasets.
Below we introduce each dataset in detail.

\paragraph{DROP} A reading comprehension benchmark to measure \textit{numerical reasoning} ability over a given passage~\cite{dua-etal-2019-drop}.
It contains three subsets of questions: \textit{span}, \textit{number}, and \textit{date}, each of which involves a lot of numerical operations. Unlike traditional reading comprehension datasets such as SQuAD \cite{rajpurkar-etal-2016-squad} where answers are always a single span from context, several answers in the \textit{span} subset of DROP contains multiple spans.
The \textit{number} and \textit{date} answers are mostly out of context and need generative-level expressiveness.

\paragraph{HotpotQA} An extractive reading comprehension dataset that requires models to perform \textit{multi-hop reasoning} over different passages~\cite{yang-etal-2018-hotpotqa}.
It contains two settings (i) \textit{Distractor}: reasoning over $2$ gold paragraphs along with $8$ similar distractor paragraphs and (ii) \textit{Full wiki}: reasoning over customized retrieval results from full Wikipedia passages.
We experiment with its distractor setting since retrieval strategy is beyond our focus in this work.

\begin{table}[t]
  \centering
  \small
  \scalebox{0.9}{
    \begin{tabular}{lcccc}
    \toprule
    \multicolumn{1}{c}{\multirow{2}[4]{*}{\bf Dataset}} & \multicolumn{2}{c}{\bf Train} & \multicolumn{2}{c}{\bf Dev}
    \\
         \cmidrule(lr){2-3}
     \cmidrule(lr){4-5}
          & {\# Questions} & {\# Docs} & {\# Questions} & {\# Docs} \\
    \midrule
    DROP & $77,409$  & $5,565$  &  $9,536$ & $582$ \\
    HotpotQA & $90,564$  & $90,564$  & $7,405$ & $7,405$\\ 
    TAT-QA & $13,215$ & $2,201$ & $1,668$  & $278$ \\
    SVAMP   & --  & --  & $726$  & $726$ \\
    EQUATE & --  & -- & $9,606$ & $9,606$ \\
    LogiQA & $6,942$  & $6,942$ & $868$ & $868$ \\
    \bottomrule
    \end{tabular}
    }
  \caption{The statistics of our experimental datasets.}
  \label{tab:dataset-data}%
\end{table}%

\paragraph{TAT-QA} A question answering benchmark to measure reasoning ability over \textit{hybrid} context, i.e., passages and tables~\cite{zhu-etal-2021-tat}.
It is curated by combing paragraphs and tables from real-world financial reports.
According to the source(s) the answers are derived from, the dataset can be divided into three subsets: \textit{Table}, \textit{Text} and \textit{Table-Text(both)}.

\paragraph{EQUATE} The first benchmark dataset to explore \textit{quantitative reasoning} under the task of natural language inference~\cite{ravichander-etal-2019-equate}.
As a test-only dataset, it requires fine-tuned models on MNLI to perform \textit{zero-shot} natural language inference tasks over quantitative statements described in (premise, hypothesis) pairs to reach final entailment decisions.

\paragraph{LogiQA} A multi-choice reading comprehension dataset that evaluates the \textit{logical reasoning} ability, whose questions are designed by domain experts~\cite{ijcai2020-501}.
It contains four types of logical reasoning, including categorical reasoning, disjunctive reasoning, conjunctive reasoning and conditional reasoning.

\paragraph{SVAMP} A challenging math word problem dataset~\cite{patel-etal-2021-nlp}. 
It is designed specifically to hack models who leverage spurious patterns to perform arithmetic operations without true understanding of context.
We only keep addition and subtraction problems in accordance with our pre-training coverage.

\subsection{Baseline Setup}

We summarize the baseline methods in short below, and refer readers to their papers for more details.
(i) On \textbf{DROP}, we include two families of models for comparison: specialized models such as NumNet(+)~\citep{ran-etal-2019-numnet}, MTMSN~\citep{hu-etal-2019-multi}, NeRd~\citep{Chen2020Neural}, QDGAT~\cite{chen-etal-2020-question} and language models such as GenBERT~\citep{geva-etal-2020-injecting} and  PReaM~\citep{DBLP:journals/corr/abs-2107-07261}.
(ii) Similarly, on \textbf{HotpotQA} (Distractor), specialized model baselines include DFGN~\citep{qiu-etal-2019-dynamically}, SAE~\citep{DBLP:conf/aaai/TuHW0HZ20}, C2F Reader~\citep{shao-etal-2020-graph} and the SOTA model HGN~\citep{fang-etal-2020-hierarchical}. The language model baselines consist of BERT~\citep{devlin-etal-2019-bert}, SpanBERT~\citep{joshi-etal-2020-spanbert} and ReasonBERT~\citep{deng2021reasonbert}.
(iii) On \textbf{TAT-QA}, we adopt the official baselines, including \tapas~\citep{herzig-etal-2020-tapas}, NumNet+ V2 and the SOTA model \tagop~\citep{zhu-etal-2021-tat}.
(iv) On \textbf{EQUATE}, we compare our methods with BERT~\citep{devlin-etal-2019-bert}, GPT~\citep{radford2019language} and Q-REAS~\citep{ravichander-etal-2019-equate}.
(v) On \textbf{LogiQA}, we compare our methods with Co-Matching Network~\citep{wang-etal-2018-co} and the SOTA model DAGN~\citep{huang-etal-2021-dagn}.

\section{Implementation Details}\label{sec:append_imp}

\subsection{\oursroberta on Different Datasets}

On \textbf{DROP}, we cast the span selection task as a sequence tagging problem following~\citet{segal-etal-2020-simple}.
On \textbf{TAT-QA}, we in-place substitute the RoBERTa-Large encoder in \tagop~\citep{zhu-etal-2021-tat} with our \oursroberta to verify its effectiveness, and keep the rest of the components unchanged.
On \textbf{HotpotQA}, we train two classifiers independently to predict the start and end positions of the answer span, as done in~\citet{devlin-etal-2019-bert}.
On \textbf{EQUATE}, we train a classifier to perform sequence classification on concatenated premise-hypothesis pairs.
Notably, we follow the official setup to train LMs on the MNLI dataset~\cite{williams-etal-2018-broad} and evaluate their zero-shot performance on EQUATE.
On \textbf{SVAMP}, the encoder-only model is not suitable since the answers are out-of-context.

\begin{table*}[t]
\centering
\scalebox{1.0}{
\begin{tabular}{lccccc}
\toprule
\textbf{Models} & \textbf{Number} & \textbf{Span} & \textbf{Spans} & \textbf{Date} & \textbf{Total}  \\
\toprule
\multicolumn{6}{c}{\textit{Previous Systems}} \\
MTMSN (BERT) & $81.1$ &  $82.8$ & $62.8$ & $69.0$  & $80.5$ \\
NumNet+ (RoBERTa) & $83.1$ &  $86.8^*$ & ${86.8}^*$ & $63.9$  & $84.4$ \\
QDGAT (RoBERTa) & ${86.2}$ &  $88.5^*$ & ${88.5}^*$ & $67.5$  & ${87.1}$ \\
GenBERT & $75.2$ & $74.5$ & $24.2$ & $56.4$ & $72.3$\\
PReasM &  $64.4$ & $86.6$ & $78.4$ & $77.7$ & $72.3$\\
\multicolumn{6}{c}{\textit{Original LMs}} \\
RoBERTa-Large & -- &  $86.4$ & $79.9$ & -- & --\\
BART-Large & $63.6$ & $79.6$ & $74.6$ & $62.1$ & $69.2$ \\ 
T5-11B & $83.2$ & ${90.2}$ & ${85.8}$ & ${84.9}$ & $85.8$\\
\multicolumn{6}{c}{\textit{\ours Models}} \\
\oursroberta & -- & $88.2$ & $83.1$ & -- & --\\
\oursbart & $78.9$ & $84.5$ & $79.6$ & $71.9$ & $80.6$\\
\ourstfive & ${85.2}$ & ${92.4}$ & ${86.6}$ & ${84.4}$ & ${87.6}$\\
\bottomrule
\end{tabular}
}
\caption{Breakdown of model \fone{} score by answer types on the dev set of DROP. Some works only report overall span type performance (marked by *), and single-span is non-separable from multi-span performance.
Bold and underlined numbers indicate the best and second-best results, respectively.
}
\label{tab:drop-by-type}    
\end{table*}

\subsection{Pre-training Details}

By default, we apply AdamW as pre-training optimizer with default scheduling parameters in fairseq.
The coefficient of weight decay is set as $0.05$ to alleviate over-fitting of pre-trained models.
Additionally, we employ fp16 to accelerate the pre-training.

\paragraph{\ours-Math} The pre-training procedure lasts for $10,000$ steps with a batch size of $512$. After the warm up in the first $2000$ steps, the learning rate arrives the peak at $3{\times}{10}^{-5}$ during pre-training.

\paragraph{\ours-Logic} The pre-training procedure lasts for $5,000$ steps with a batch size of $512$. After the warm up in the first $1000$ steps, the learning rate arrives the peak at $3{\times}{10}^{-5}$ during pre-training.

\paragraph{\ours-SQL} For \oursbart and \oursroberta, the pre-training procedure lasts for $50,000$ steps with a batch size of $512$. After the warm up in the first $5000$ steps, the learning rate arrives the peak at $3{\times}{10}^{-5}$ during pre-training. To save memory, each example in the pre-training corpus could at most contains $512$ tokens.
For \ourstfive, the pre-training procedure lasts for $20,000$ steps with a batch size of $512$. After the warm up in the first $2000$ steps, the learning rate arrives the peak at $1{\times}{10}^{-5}$ during pre-training. The maximum input length in each example is truncated to $384$ tokens to increase the batch size.

\begin{table*}[ht]
    \small
    \centering
    \begin{tabular}{lcccccc}
    \toprule
    \textbf{Models} &  \textbf{RTE-Q} &  \textbf{NewsNLI} &  \textbf{RedditNLI} &  \textbf{NR ST} &  \textbf{AWPNLI} &  \textbf{Average}  \\
    \midrule
    \multicolumn{7}{c}{\textit{Previous Systems}} \\
    MAJ & $57.8$ & $50.7$ & $58.4$ & $33.3$ & $50.0$ & $50.4$ \\
    BERT  & $57.2$ & $72.8$ & $49.6$ & $36.9$ & $42.2$ & $51.8$   \\
    GPT  & $68.1$ & $72.2$ & $52.4$ & $36.4$ & $50.0$ & $55.8$  \\
    Q-REAS & $56.6$ & $61.1$ & $50.8$ & $63.3$ & ${71.5}$ & $60.7$ \\
    \multicolumn{7}{c}{\textit{Original LMs}} \\
    BART-Large & $68.1$ & ${76.2}$ & $65.0$ & $53.7$ & $49.7$ & $62.6$   \\
    RoBERTa-Large & $69.3$ & ${75.5}$ & ${65.6}$ & $60.1$ & ${50.7}$ & $64.2$   \\
    \multicolumn{7}{c}{\textit{\ours Models}} \\
    \oursbart & ${72.3}$ & $75.2$ & $64.8$ & ${70.7}$ & $49.5$ & ${66.5}$   \\
    \oursroberta & ${75.3}$ & ${75.5}$ & ${68.1}$ & ${69.2}$ & $50.5$ & ${67.5}$   \\
    \bottomrule
    \end{tabular}
    \caption{The EM performance of different models on all subsets of the EQUATE benchmark. Bold and underlined numbers indicate the best and second-best results, respectively.}
    \label{tab:equate}   
\end{table*}

\begin{table*}[ht]
\centering
\small
\scalebox{1.0}{
\begin{tabular}{lcccc}
\toprule
&  \textbf{Table} &  \textbf{Text} &  \textbf{Table-Text} &  \textbf{Total} \\
\cmidrule{2-5} & EM\,\,/\,\,\fone{} & EM\,\,/\,\,\fone{} & EM\,\,/\,\,\fone{} & EM\,\,/\,\,\fone{}
\\
\midrule
Arithmetic & $50.1$\,\,/\,\,$50.1$ & $43.8$\,\,/\,\,$50.0$ & $55.6$\,\,/\,\,$55.6$ & $51.5$\,\,/\,\,$51.5$ \\
Counting & $66.7$\,\,/\,\,$66.7$ &  --\,\,\,/\,\,\,-- & $90.0$\,\,/\,\,$90.0$ & $81.3$\,\,/\,\,$81.3$\\
Spans & $67.4$\,\,/\,\,$80.6$ & $54.2$\,\,/\,\,$80.8$ & $79.2$\,\,/\,\,$84.8$ & $71.4$\,\,/\,\,$82.6$\\
Span &  $68.4$\,\,/\,\,$68.4$ & $51.2$\,\,/\,\,$76.0$ & $76.2$\,\,/\,\,$77.8$ & $61.9$\,\,/\,\,$74.6$\\
Total & $56.5\,\,/\,\,58.0$ & $51.1\,\,/\,\,75.0$ & $69.0\,\,/\,\,70.7$ & $59.1$\,\,/\,\,$65.9$ \\
\bottomrule
\end{tabular}}
\caption{The EM performance of \tagop (\oursroberta) with respect to answer types and sources on the dev set of TAT-QA.
}
\label{tab:tatqa}    
\end{table*}

\subsection{Fintuning Details}

We implement our models based on transformers~\cite{wolf-etal-2020-transformers}, fairseq~\cite{ott-etal-2019-fairseq} and DeepSpeed~\footnote{\url{http://github.com/microsoft/DeepSpeed}}.

\paragraph{Passage Retrieval in HotpotQA} Since the total length of the original passages in HotpotQA is too long to fit into memory, we train a classifier to filter out top-$3$ passages, as done in previous work~\citep{deng2021reasonbert}.
Specifically, a RoBERTa-Large model is fine-tuned to discriminate if an input passage is required to answer the question. 
The Hits@$3$ score of the classifier on HotpotQA is $97.2\%$.

\paragraph{Numerical Design in DROP and SVAMP}

As noticed by previous works, sub-word tokenization methods such as byte pair encoding \cite{DBLP:journals/corr/SennrichHB15} potentially undermines the arithmetic ability of models.
Instead, the character-level number representation is argued to be a more effective alleviation~\cite{wallace-etal-2019-nlp}.
Additionally, the reverse decoding of numbers is proposed as a better way of modelling arithmetic carry~\citep{geva-etal-2020-injecting}. 
Therefore, we employ these design strategies on DROP and SVAMP.

\subsection{Fine-tuning Hyperpameters}

By default, we apply AdamW as fine-tuning optimizer with default scheduling parameters on all datasets. To ensure statistical significance, all fine-tuning procedures are run with three random seeds, except for T5-$11$B and \ourstfive due to the limit of computation budgets.

\paragraph{DROP}
\oursroberta and RoBERTa-Large are trained with the subset of questions marked as ``span'' from the DROP dataset.t
Since a gold answer may occur multiple times in the passage, we optimize over the sum of negative log probability for all possibly-correct \texttt{IO} sequences where each one of gold answers is included at least once, as done in \citet{segal-etal-2020-simple}.
The fine-tuning procedure runs up to $25,000$ steps with a batch size of $64$, with the learning rate of $7.5{\times}{10}^{-6}$.
As for BART-Large (and \oursbart, \oursbartmath, the same below) and T5-$11$B (and \ourstfive, the same below), they are trained with the whole DROP dataset.
For BART-Large, the fine-tuning procedure runs up to $20,000$ steps with a batch size as $128$ and a learning rate as $3{\times}{10}^{-5}$.
For T5-$11$B, due to the computational budget, the fine-tuning procedure only lasts for $10,000$ steps with a batch size of $32$, and the learning rate is $1{\times}{10}^{-5}$.

\paragraph{TAT-QA}

In the experiment of TAT-QA, we employ the official implementation and the default hyperparameters provided in \tagop~\footnote{https://github.com/NExTplusplus/TAT-QA}.
The fine-tuning procedure runs up to $50$ epochs with a batch size of $48$.
For modules introduced in \tagop, the learning rate is set as $5{\times}{10}^{-4}$, while for RoBERTa-Large (and \oursroberta), the learning rate is set as $1.5{\times}{10}^{-5}$.

\paragraph{HotpotQA}

The fine-tuning procedure runs up to $30,000$ steps with a batch size of $64$. The learning rate is $1{\times}{10}^{-5}$. Overlong inputs are truncated to $512$ tokens for both RoBERTa-Large (and \oursroberta), T5-$11$B (and \ourstfive) and BART-Large (and \oursbart). 

\paragraph{EQUATE}
The fine-tuning procedure runs up to $20,000$ steps on MNLI with a batch size of $128$ for both RoBERTa-Large (and \oursroberta) and BART-Large (and \oursbart), with learning rate is $1{\times}{10}^{-5}$. After fine-tuning, models are directly evaluated on EQUATE.

\paragraph{LogiQA}

In the experiment of LogiQA, we employ the open-source implementation and the default hyperparameters provided in ReClor~\footnote{https://github.com/yuweihao/reclor} \citep{yu2020reclor} to fine-tune RoBERTa-Large (and \oursroberta).
The fine-tuning procedure runs up to $10$ epochs with a batch size of $24$.
The learning rate is set as $1{\times}{10}^{-5}$.

\section{Fine-grained Analysis}\label{sec:fine_analysis}

\paragraph{DROP} In \tabref{drop-by-type} we report model \fone{} scores by question type on DROP. Comparing three \ours pre-trained models with their vanilla versions, we observe that: (i) \oursbart outperforms the vanilla BART-large with a wide margin in all types of questions, i.e. \textit{number} ($15.3\%$), \textit{date} ($9.8 \%$), \textit{span} (around $5\%$). (ii) \oursroberta only deals with span selection questions, and obtain $1.9\%$, $3.2\%$ gain on \textit{span, spans} questions, respectively.
(iii) For the giant \ourstfive, we also observe $2\%$ improvement on \textit{number} questions, $2.2\%$ on \textit{span} and $0.8\%$ on \textit{spans} questions. These model-agnostic performance boost on DROP reveals the extra numerical reasoning knowledge models learned from SQL program executors.

\paragraph{EQUATE}
\tabref{equate} presents performance break-down by subsets of EQUATE \cite{ravichander-etal-2019-equate}, where we compare \oursbart and \oursroberta with their vanilla versions and previous baselines. For both models, we observe around $10\%$ acc improvement on the \textit{NR ST} subset, where \textbf{numerical comparison and quantifiers} are especially emphasized. Stable performance improvement was also observed in both pre-trained models on  the \textit{RTE-Q} subset, where \textbf{arithmetics and ranges} are primary focus. Interestingly, \oursroberta alone demonstrate improvement on \textit{RedditNLI} (emphasizes approximation and verbal quantitative reasoning) subset. Performance on other subsets are approximately comparable between \ours pre-trained models and vanilla models, suggesting that \ours does not harm intrinsic abilities of language models.

\paragraph{TAT-QA}

\tabref{tatqa} shows the detailed experimental results of \tagop (\oursroberta). 
Considering that the pre-training of \oursroberta is only performed on table-like texts (i.e., the flatten sequence of databases), it is highly non-trivial for our model to generalize to such a hybrid scenario containing both tables and passages, again illustrating the transferability of reasoning capabilities.

\section{NL Understanding Performance}\label{sec:nl_understnad}

\begin{table}[t]
  \centering
  \small
  \scalebox{0.9}{
    \begin{tabular}{lcccc}
    \toprule
    \multicolumn{1}{c}{\multirow{2}[4]{*}{\bf Dataset}} & \multicolumn{2}{c}{\bf Train} & \multicolumn{2}{c}{\bf Dev}
    \\
         \cmidrule(lr){2-3}
     \cmidrule(lr){4-5}
          & {\# Questions} & {\# Docs} & {\# Questions} & {\# Docs} \\
    \midrule
    SQuAD & $77,409$  & $5,565$  &  $9,536$ & $582$ \\
    MNLI & $392,702$  & $392,702$  & $9,815$ & $9,815$\\ 
    QuoRef & $19,399$ & $3,771$ & $2,418$  & $454$ \\
    \bottomrule
    \end{tabular}
    }
  \caption{\ours on NL understanding experiment dataset statistics.}
  \label{tab:mrc-stat}%
\end{table}%

\paragraph{Dataset Details} We fine-tune \oursroberta on (i) SQuAD v1.0: \citep{rajpurkar-etal-2016-squad}: one of the most classical single-span selection RC benchmarks measuring understanding over natural language context; (ii) MNLI \citep{williams-etal-2018-broad}: a large-scale NLI dataset measuring cross-domain and cross-genre generalization of NLU. Notably, our model is evaluated on the \textit{matched} setting for the purpose of simplicity. (iii) QuoRef \cite{dasigi-etal-2019-quoref}: A Wikipedia-based multi-span selection RC benchmark with a special emphasis on coreference resolution. All dataset Statistics are shown in \tabref{mrc-stat}.

\paragraph{Implementation Details}
(i) On SQuAD, we cast the span selection task as a sequence tagging problem following~\citet{segal-etal-2020-simple}. 
(ii) On MNLI-matched, we train both models to perform sequence classification on concatenated premise-hypothesis pairs.
(iii) On {Quoref}, we cast the span(s) selection task as an \texttt{IO} sequence tagging problem following~\citet{segal-etal-2020-simple}.

\end{document}